\documentclass{article}

\usepackage{arxiv}

\usepackage[utf8]{inputenc} 
\usepackage[T1]{fontenc} 
\usepackage{amsmath,amssymb,amsfonts}%
\usepackage{hyperref}       
\usepackage{url}            
\usepackage{booktabs}       
\usepackage{amsfonts}       
\usepackage{nicefrac}       
\usepackage{microtype}      
\usepackage{lipsum}
\usepackage{graphicx}
\usepackage{algorithm}
\usepackage{algorithmicx}
\usepackage{algpseudocode}
\usepackage{listings}
\usepackage{array}
\graphicspath{ {./images/} }

\title{Autonomous Vehicle Decision-Making Framework for Considering Malicious Behavior at Unsignalized Intersections}

\author{
 Qing Li \\
  Shandong University of Science and Technology\\
  Qingdao, Shandong 266590 \\
  \texttt{liqing@sdust.edu.cn} \\
   \And
 Jinxing Hua \\
  Shandong University of Science and Technology\\
  Qingdao, Shandong 266590 \\
  \texttt{huajinxing0821@163.com} \\
  \And
 Qiuxia Sun \\
  Shandong University of Science and Technology\\
  Qingdao, Shandong 266590 \\
  \texttt{qiuxiasun@sdust.edu.cn} \\
}

\begin{document}
\maketitle
\begin{abstract}
In this paper, we propose a Q-learning based decision-making framework to improve the safety and efficiency of Autonomous Vehicles when they encounter other maliciously behaving vehicles while passing through unsignalized intersections. In Autonomous Vehicles, conventional reward signals are set as regular rewards regarding feedback factors such as safety and efficiency. In this paper, safety gains are modulated by variable weighting parameters to ensure that safety can be emphasized more in emergency situations. The framework proposed in this paper introduces first-order theory of mind inferences on top of conventional rewards, using first-order beliefs as additional reward signals. The decision framework enables Autonomous Vehicles to make informed decisions when encountering vehicles with potentially malicious behaviors at unsignalized intersections, thereby improving the overall safety and efficiency of Autonomous Vehicle transportation systems. In order to verify the performance of the decision framework, this paper uses Prescan/Simulink co- simulations for simulation, and the results show that the performance of the decision framework can meet the set requirements. 
\end{abstract}

\keywords{Autonomous Vehicles \and Q-learning \and Decision-making \and Unsignalized Intersection}

\section{Introduction}
With the continuous advancement of autonomous driving technology, the expectation of achieving a safer, more efficient, and more reliable road transportation system is increasing. Autonomous vehicles are expected to have a lower accident rate compared to human drivers. However, in the future, autonomous vehicles will share the road with ordinary vehicles. Achieving safety in mixed and complex traffic scenarios is challenging. To ensure the overall safety and efficiency of the traffic system, the decision-making capability of autonomous vehicles is crucial. They need to make accurate and reliable decisions in complex traffic environments.

Research on unsignalized intersections has become a focus due to their high accident rates. In the EU27, intersection accidents account for 43\% of road injury accidents\cite{simon2009intersection}. In the United States, about 40\% of the 5,811,000 crashes in 2008 were related to intersections\cite{choi2010crash}. Unsignalized intersections, lacking signalized scheduling, offer higher maneuverability and flexibility for vehicle traffic compared to signalized intersections. This increases the likelihood of malicious behavior by vehicles, making unsignalized intersections more prone to collisions\cite{chen2022cooperation} and reducing the safety of the transportation system. \emph{The Global Status Report on Road Safety 2023} shows that only one-fifth of the world's roads meet basic safety requirements. Therefore, there is an urgent need to develop a safe and effective decision-making model for risk avoidance at unsignalized intersections.

Currently, the key challenge in decision-making for autonomous vehicles is dealing with the uncertainty of traffic participants, especially vehicles exhibiting malicious behavior. Traditional research has approached this challenge by transforming it into an optimization problem. Xu combines Monte Carlo Tree Search (MCTS)\cite{li2006cooperative} with heuristic rules to find a nearly globally optimal sequence of passes for autonomous vehicles. However, this approach struggles with more complex road conditions\cite{xu2019cooperative}. Moreau proposes an optimization method using Bessel curves\cite{costanzi2016generic}\cite{nayl2013online} and the Newton-Raphson gradient method to handle unexpected events or obstacles\cite{moreau2019reactive}. This method allows vehicles to safely navigate complex scenarios, such as intersections with obstacles. However, it overlooks the continuity of interactions between vehicles, which may prevent them from making the right judgments in emergency situations.

With the advancement of artificial intelligence, data-driven decision-making has become increasingly prevalent in the field of autonomous vehicles. Machine learning enables the extraction of key insights from vast amounts of sensor data, enhancing the accuracy and reliability of decisions. To address uncertainty, \cite{wei2011point} modeled the decision-making process of autonomous vehicles as a Markov Decision Process (MDP)\cite{thrun2002probabilistic} and introduced a QMDP-based\cite{murphy2000survey} behavioral approach for single-lane scenarios. However, this method's state space is limited to a single lane and is inadequate for complex traffic environments such as intersections. To overcome this limitation, \cite{brechtel2013solving} formulated the driving task as a continuous Partially Observable Markov Decision Process (POMDP)\cite{sondik1971optimal}\cite{kaelbling1998planning} and developed a solver to derive near-optimal policies, addressing decision-making challenges in uncertain driving environments\cite{brechtel2014probabilistic}. Nonetheless, ensuring the safety and efficiency of autonomous vehicle decision-making remains challenging when other traffic participants behave unexpectedly.

Reinforcement learning (RL), known for its efficiency and scalability, has become a key tool for training autonomous vehicle algorithms\cite{martin2016vision,chakraborty2010extended,ccetin2014path,zhang2008adaptive,mazouchi2022automating}. Mazouchi proposed a risk-averse Q-learning planner\cite{mihatsch2002risk,sutton1999reinforcement} that uses teaser information and reward functions to generate samples without directly interacting with the environment, thereby learning optimal planning strategies for risk aversion. However, this approach requires data collection for each new MDP, which is time-consuming and impractical\cite{mazouchi2022automating}. In a separate study, Liu introduced a reinforcement learning framework, DPL, designed to enhance decision-making and social behavior in autonomous vehicles by incorporating driving priors and socially coordinated awareness. Despite these advancements, the DPL approach can be egocentric, causing autonomous vehicles to overlook interactions with other traffic participants\cite{liu2023towards}. While machine learning excels in handling uncertainty and making real-time or near-real-time decisions, its complex structures and parameters often make the decision-making process difficult to interpret and understand.

Compared to machine learning, game theory offers a clearer framework for describing interactions and decision-making among vehicles, making the results more interpretable and understandable\cite{hang2020human,li2016hierarchical,lopez2022game,rahmati2017towards,nan2022intention,tian2020game}. Vehicles optimize their strategies by considering the potential actions of other traffic participants. Pruekprasert proposed a game-theoretic decision-making approach that determines vehicle priorities while accounting for the aggressive and unpredictable behavior of malicious vehicles to simulate real traffic environments. However, the approach's definition of a "demonic vehicle," which always prioritizes itself without considering others, results in relatively fixed priority orders\cite{pruekprasert2019decision}. Liu introduced a three-tier decision-making framework consisting of an action space filter, a normal-form two-player game, and a safety check, with neural network tuning to handle various driving styles flexibly\cite{liu2022three}. Lu proposed an innovative framework based on a non-cooperative dynamic game algorithm, reducing game complexity by setting an entry threshold. This framework considers driving efficiency, safety, and comfort, and its effectiveness was validated in scenarios involving four different aggressive vehicles\cite{lu2023game}. While these game-theoretic approaches account for different driving styles, doing so may increase model complexity, requiring more parameters, strategies, and logic, thus complicating development and implementation. Additionally, building accurate models of driving styles demands substantial data and complex modeling techniques, which may not be feasible in real-world scenarios. To address these challenges, we propose a novel decision-making framework.

\begin{figure*}[t]
\centering
\includegraphics[width=0.8\textwidth, height=0.4\textheight]{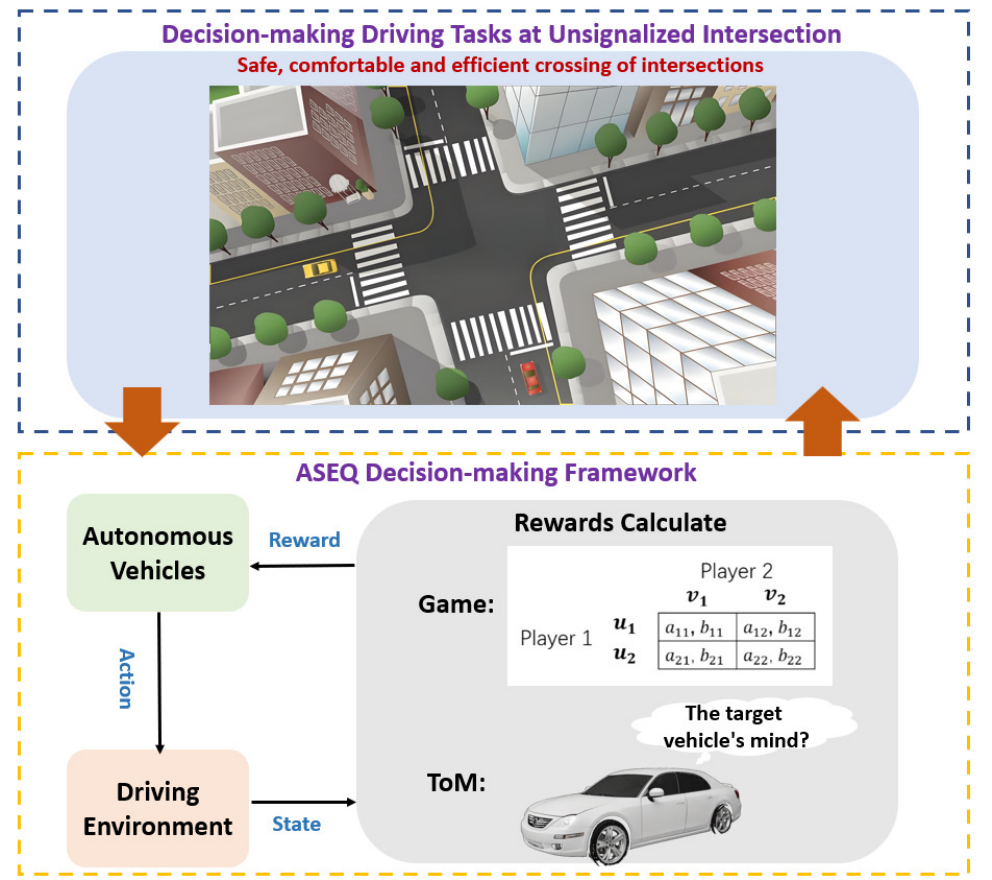}
\caption{Decision-Making Framework for Autonomous Vehicles at Unsignalized Intersections}
\label{intro}
\end{figure*}

The main contributions of this paper are: (1) We design an Adaptive Safety-Enhanced Q-Learning Framework (ASEQ) for unsignalized intersections, as shown in Fig.\ref{intro}. (2) We integrate both emergency and non-emergency risk avoidance into the game-theoretic decision-making process. By introducing Time-to-Collision (TTC) as a key metric and designing variable weight parameters, we dynamically adjust these weights based on changes in TTC to balance safety, comfort, and efficiency in driving decisions. By increasing the emphasis on safety, the framework proactively reduces the risk of potential collisions. In emergency situations, safety is prioritized to minimize the likelihood of collisions during avoidance maneuvers. (3) We employ a first-order Theory of Mind (ToM) inference approach to better understand the motivations behind the target vehicle's behavior, serving as an alternative to traditional studies on aggressive behavior. This approach allows us to infer the target vehicle's intrinsic motivations and assess the probability of malicious actions, providing critical insights for the decision-making process.

The rest of the paper consists of the following sections: Section II details the design of the decision, discusses the key factors to be considered in the decision-making process, and proposes a Q-learning-based decision-making model. Section III provides the researcher with an understanding of the challenges and risks that may arise during the decision-making process by simulating malicious behavior. In Section IV, a simulation study is conducted to validate the performance of the designed algorithm. The last section summarizes and concludes the findings by discussing the strengths, limitations, and potential directions for future research on the algorithm.

\section{Decision-making design}\label{sec2}

\begin{figure}[t]
\centerline{\includegraphics[width=3.5in]{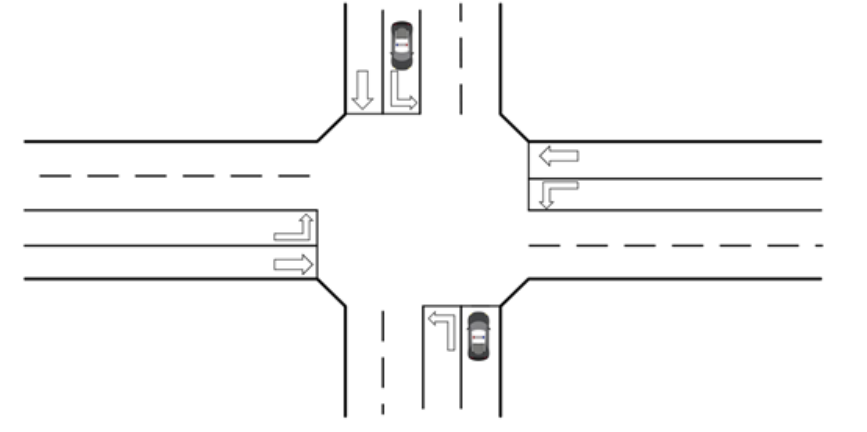}}
\caption{Schematic diagram of intersection\label{Fig1}}
\end{figure}

We consider an autonomous vehicle at a four-way unsignalized intersection, as illustrated in Fig.\ref{Fig1}. The autonomous vehicle is traveling in the right lane (i.e., the straight lane), while an unknown left-turning vehicle is approaching from the opposite direction. The autonomous vehicle's decision-making is influenced by vehicles entering or currently within the intersection, but not by those that have already exited. Fig.\ref{Fig1} depicts a typical scenario where Vehicle 1, the target vehicle, may exhibit potentially malicious behavior, and Vehicle 2 is the autonomous vehicle. The potential conflict zone lies at the intersection of their future trajectories.

The relevant assumptions are as follows: 1) Vehicles share status information (e.g., speed, position) and other necessary data via Vehicle-to-Everything (V2X) technology, enabling communication between vehicles and infrastructure, where V represents the vehicle and X represents various interacting entities (e.g., infrastructure, other vehicles, pedestrians). Each vehicle is aware of the payoffs of other vehicles; 2) Each vehicle follows a pre-planned path through the intersection and adjusts its speed to avoid conflicts with other vehicles; 3) Left-turn maneuvers are permitted only at the intersection; and 4) Lane changes are not allowed while traveling within the intersection.

\begin{algorithm}
\caption{ASEQ Decision-Making Framework for Autonomous Vehicles at Unsignalized Intersections}
\label{alg:aseq_learning}
\begin{algorithmic}[1]
\State Initialize Q-table $Q(s, a)$ with zeros
\State Set learning rate $\alpha$, discount factor $\gamma$, exploration rate $\epsilon$
\State Set reward weight parameters $W_1$ and $W_2$
\For{each episode}
    \State Initialize state $s$
    \While{state $s$ is not terminal}
        \State Select action $a$ using $\epsilon$-greedy policy
        \State Take action $a$, observe reward $r$ and next state $s'$
        
        \State \textbf{First-order ToM Inference:}
        \State Compute ToM reward $R_{\text{ToM}}$
        
        \State \textbf{Safety, Efficiency, and Comfort Payoff Calculation:}
        \State Compute $W_s(TTC)$, $f_{i,s}(t)$, $f_{i,e}(t)$, and $f_{i,c}(t)$
        
        \State \textbf{Total Payoff Calculation:}
        \State $U_i(t) = W_s(TTC) \cdot f_{i,s}(t) + W_e \cdot f_{i,e}(t) + W_c \cdot f_{i,c}(t)$
        
        \State \textbf{Game Reward Calculation:}
        \State Compute strategy combination $\pi$ and payoff $U_i(t)$
        \State Find Nash equilibrium $\pi^*_{\text{Nash}}$ using:
        \State $\pi^*_{\text{Nash}} = \arg\min_{\pi_{\text{Nash}} \in \Pi_{\text{Nash}}} \|\pi - \pi_{\text{Nash}}\|$
        \State Compute game reward $R_{\text{Game}}$ using:
        \State $R_{\text{Game}} = \frac{1}{1 + e^{-\delta (\pi - \pi^*_{\text{Nash}})}}$
        
        \State \textbf{Total Reward Calculation:}
        \State $R = W_1 \cdot R_{\text{ToM}} + W_2 \cdot R_{\text{Game}}$
        
        \State \textbf{Q-value Update:}
        \State $Q(s, a) \leftarrow Q(s, a) + \alpha \left( R + \gamma \max_{a'} Q(s', a') - Q(s, a) \right)$
        \State Set state $s \leftarrow s'$
    \EndWhile
    \State Decay exploration rate $\epsilon$
\EndFor
\end{algorithmic}
\end{algorithm}

\subsection{Definition of vehicles}\label{sub_1}
In this paper, a mathematical model of a discrete-time system is utilized to define the changes in the state of a vehicle, 

\begin{equation}
\label{eq1}
s_{i,t+1}=f\left(s_{i,t},A_{i,t})\right.
\end{equation}

where $s_{i,t}$ is the kinematic state of vehicle i at time t.$A_{i,t}$ is the action that vehicle i decides to take at moment t.We define that,

\begin{equation}
\label{eq2}
s_{i,t}=\left(x_{i,t},y_{i,t},v_{x_{i,t}},v_{y_{i,t}},a_{i,t+1})\right.
\end{equation}

where $x_{i,t}$,$y_{i,t}$ is the position of vehicle i at time t,$v_{x_{i,t}}$,$v_{y_{i,t}}$ is the velocity of vehicle i in the x-axis direction and y-axis direction at moment t. The navigation path of the autonomous vehicle is a fixed path, considering that the goal of the autonomous vehicle is fixed, i.e., to cross an unsignalized intersection. Therefore, the autonomous vehicle can only control its acceleration along the navigation path at each step. Such a state definition provides a comprehensive description of the vehicle's kinematic state and provides enough information to infer the vehicle's behavior and decisions. By updating the state variables, we can model the vehicle's state changes over discrete time steps and make decisions based on the current state at each time step.

\subsection{State space}
In the actual driving operation of crossing an unsignalized intersection, the driver needs to make a decision accordingly based on the position and speed of the target vehicle as well as the autonomous vehicle. Therefore, according to the vehicle state definition, this paper sets the state space S as the vehicular kinematic state of the target vehicle as well as the autonomous vehicle, so that the state $S_t$ can be expressed as,

\begin{equation}
\label{eq3}
S_{t}=\left(s_{1,t},s_{2,t})\right.
\end{equation}

where $i\in(1,2)$ for vehicles in unsignalized intersections, we define Vehicle 1 as a target vehicle, i.e., a vehicle that may commit malicious acts, and Vehicle 2 as an autonomous vehicle.

\subsection{Action space}
From \ref{sub_1} we assume that the autonomous vehicle can only control its acceleration along the navigation path at each time step, so for an autonomous vehicle the action space can be defined as, 

\begin{equation}
\begin{aligned}
A_{i,t} &= \left(a_{i,t+1}^1, a_{i,t+1}^2, ..., a_{i,t+1}^j, ..., a_{i,t+1}^n\right), a_{i,t+1}^j \in (a_{\text{min}}, a_{\text{max}}).
\end{aligned}
\label{eq4}
\end{equation}

where $a_{i,t+1}^j$ is the acceleration strategy j of the autonomous vehicle from moment t to moment t+1. $a_min$ is the minimum acceleration of the autonomous vehicle (i.e., the maximum braking acceleration),$a_max$ is the maximum acceleration of the autonomous vehicle, this means that the action space of an autonomous vehicle is only related to the relevant configuration of the autonomous vehicle. 

In many studies, although the determination of continuous action space is also carried out by restricting the reasonable range of relevant variables, the calculation of continuous action space often requires more computational cost while more refined decision judgment, which may lead to the inability to make timely and correct judgments in practical applications, and the difference between the fineness of continuous and discrete action space is significantly reduced in the case of small decision interval. So in this paper, we use discrete action space and make decisions only and driving paths are planned by the rest of the modules.

\subsection{Reward}\label{2.D}
In the field of reinforcement learning, the reward function plays a crucial role in not only guiding the behavioral intentions of an agent but also determining which strategy the agent will adopt for learning. A well-designed reward function can significantly improve the convergence speed of an algorithm and ensure that the decision-making process of a reinforcement learning agent is both safe and efficient. In real-world driving scenarios, the decision-making module needs to satisfy multiple requirements, including avoiding collisions with other vehicles, maximizing driving speed while ensuring safety, and reducing unnecessary acceleration and deceleration. In view of this, the reward function structure proposed in this study consists of two main parts, aiming to comprehensively consider multiple driving requirements in order to achieve the optimal behavioral strategies.

Most of the current reward functions about guiding agents to safely pass through complex traffic scenarios give a negative reward for behaviors that may be dangerous. In traffic scenarios, the behaviors and decisions of an agent are often influenced by other traffic participants; however, this reward mechanism usually focuses only on the behavior of the agent itself, without considering the behaviors and strategies of other participants. This may lead to the inability of the agent to accurately model and predict the behaviors of other participants and make decisions accordingly; and it is not in line with the principle of safety prioritization, which may be ignored by simply giving positive rewards for non-collision actions. In some cases, although an action does not lead to a collision, it may still be potentially dangerous or violate traffic rules. Therefore, this paper incorporates game theory into the reward function, where conflict and competition exist in traffic scenarios, and different participants may pursue their own interests. Game theory provides a way to analyze the optimal strategies in conflict and competition situations.

\subsection*{Gaming Rewards}\label{Game}
In this study, we design a reward function that integrates safety, efficiency, and comfort with the aim of guiding agents to adopt safe driving behaviors and efficiently pass through unsignalized intersections in traffic scenarios. Specifically, if a decision on an autonomous vehicle may increase the risk of collision, the safety reward function will reduce the corresponding reward to discourage the choice of that decision. Similarly, if a decision may result in a longer time to pass through an intersection, the efficiency reward function will also reduce the reward to prevent the agent from taking such action. In addition, if a decision may reduce the comfort of the driver or passenger, the comfort reward function will similarly reduce the corresponding reward to ensure that the agent avoids choosing such an action. Therefore, we design the payoff function $U_i(t)$ of an autonomous vehicle as,

\begin{equation}
U_i\left(t\right)=W_s\cdot f_{i,s}\left(t\right)+W_e\cdot f_{i,e}\left(t\right)+W_c\cdot f_{i,c}\left(t\right)
\label{eq5}
\end{equation}

where $f_{i,s}$,$f_{i,e}$,$f_{i,c}$ are the safety, efficiency, and comfort payoff functions of vehicle i, and $W_s$,$W_e$,$W_s$ are the weights of each payoff function and $W_s+W_e+W_c=1$, and the design of the safety, efficiency, and comfort payoff functions are considered next. 

\subsubsection{Safety payoff function}

The safety reward function evaluates the degree of safety of a decision, and in order to achieve safety, this study focuses on the importance of obeying traffic rules and avoiding potential collisions or dangerous behaviors. Therefore, the violation of traffic rules or the occurrence of collisions is penalized by giving lower rewards in unsignalized intersections.

First of all, due to the lack of signals to control intersections, it is necessary to consider the situation where a vehicle moves slowly or stops at an intersection due to an unexpected situation in order to avoid other traffic participants who suddenly enter the intersection; However, vehicles moving slowly or stopping at intersections are unwanted by most of the traffic participants, which not only affects the efficiency of traffic flow, but also leads to a much higher risk of collision. Enhancement, we introduce a distance-based exponential decay function $E(t)$ that,

\begin{equation}
E\left(t\right)=\exp\left({-\beta\cdot d\left(x_{i,t},y_{i,t},x_c,y_c\right)}\right)
\label{eq6}
\end{equation}

where $(x_c,y_c)$ are the coordinates of the center of the intersection, and $\beta$ is a positive attenuation coefficient used to adjust the rate at which the gains fall. This coefficient can be adjusted according to the actual traffic environment and safety requirements $d\left(x_{i,t},y_{i,t},x_c,y_c\right)$ is the Euclidean distance from vehicle i to the center of the intersection as follows,

\begin{equation}
d\left(x_{i,t},y_{i,t},x_c,y_c\right)=\sqrt{\left(x_{i,t}-x_c\right)^2+\left(y_{i,t}-y_c\right)^2}.
\label{eq7}
\end{equation}

Second, the lower the expected collision time of a vehicle in an unsignalized intersection, the more dangerous it tends to be, and conversely, the higher the expected collision time, the safer it tends to be, so we introduce the collision time, TTC,

\begin{equation}
TTC = \frac{\vec{r} \cdot \vec{v}_{\text{rel}}}{\|\vec{v}_{\text{rel}}\|^2}
\label{eq8}
\end{equation}

where $\vec{v}_{\text{rel}} = (v_{x_{1,t}} - v_{x_{2,t}}, v_{y_{1,t}} - v_{y_{2,t}})$ is the relative velocity vector, which represents the relative difference in the speed of the two vehicles. $\vec{r} = (x_{1,t} - x_{2,t}, y_{1,t} - y_{2,t})$ is the position vector, which represents the relative position between the two vehicles. $(x_{1,t}, y_{1,t})$ are the position coordinates of vehicle 1 and $(x_{2,t}, y_{2,t})$ are the position coordinates of vehicle 2.

Next, the distance-based exponential decay function $E(t)$ and the collision time TTC are used to define the safety payoff function $f_{i,s}(t)$ that,

\begin{equation}
f_{i,s}(t) = 
\begin{cases} 
E(t + 1), & \text{if } v_{i,t+1} \leq v_{-t} \\
1 - e^{-k_1(TTC - TTC_{\text{min}})}, & \text{if } v_{i,t+1} > v_{-t}
\end{cases}
\label{eq9}
\end{equation}

where $v_{-t}$ is the threshold speed, $TTC_{\text{min}}$ is the value of TTC at the maximum speed allowed by local law, and $k_1$ is a positive constant of proportionality to control the sensitivity of the safety payoffs to changes in TTC. That is, in the next decision of vehicle $i$, if the speed value does not reach the threshold speed, i.e., if it chooses to carry out a slow travel or stop, then its safety payoff value will be calculated based on the speed value reaches the threshold speed, then the safety payoff is calculated based on the TTC value, the closer the TTC value is to $TTC_{\text{min}}$, then the lower the safety payoff value obtained by vehicle $i$. On the contrary, the higher the TTC value is, the higher the safety payoff value obtained by vehicle $i$.

\subsubsection{Efficiency payoff function}

In real-world driving, agents need to be able to achieve high efficiencies in the transportation system, and drivers often choose to complete the driving task with minimal cost in driving time while ensuring safety. For this reason, we can set positive payoffs to encourage the agent to complete the task quickly, Such a design will motivate the agent to contribute positively in terms of traffic smoothness and efficiency. In this paper, we consider an exponential decay function $f_{i,e}(t)$ that,

\begin{equation}
f_{i,e}(t) = e^{-\frac{k_2(t - t_{\text{min}})}{t_{\text{min}}}}
\label{eq10}
\end{equation}

where $t$ is the time required for vehicle $i$ to travel the remaining distance through the intersection at the post-decision speed, $t_{\text{min}}$ is the time required to travel the remaining distance through the intersection at the maximum hourly speed required to travel through the intersection according to the local law, and $k_2$ is a positive constant of proportionality to control the sensitivity of the efficiency payoffs to changes in the passage time. $t$ is small to provide high-efficiency payoffs reflecting near-ideal efficiency. As it increases, the efficiency payoff decreases rapidly, consistent with the effect of time delay on efficiency in real-world driving.

\subsubsection{Comfort payoff function}

Frequent or excessive changes in acceleration and deceleration usually make the driving process unsmooth, thus affecting the driving experience. In this paper, we hope to make driving decisions that reduce this situation in order to avoid passenger discomfort during the driving process. The comfort reward function $f_{i,c}(t)$ should be designed in such a way that it encourages the vehicle to make smooth acceleration and deceleration, and in this paper, based on the results in [32] on the comfort reward function $f_a$,
\begin{equation}
f_a(t) = 
\begin{cases} 
1 - \left| \frac{a_{i,t+1}^j - a_{i,t}}{a_{\text{max}} - \bar{a}} \right|, & \text{if } a_i^{n-1} \leq \bar{a} \\
1 - \left| \frac{a_{i,t+1}^j - a_{i,t}}{\bar{a} - a_{\text{min}}} \right|, & \text{if } a_i^{n-1} > \bar{a}\\
\bar{a} = \frac{a_{\text{max}} + a_{\text{min}}}{2}
\end{cases}
\label{eq11}
\end{equation}
which is optimally modified to obtain $f_{i,c}(t)$,
\begin{equation}
f_{i,c}(t) = 1 - \frac{k_3 \left| a_{i,t+1}^j - a_{i,t} \right|}{a_{\text{max}} - a_{\text{min}}}
\label{eq12}
\end{equation}

where $k_3$ a positive constant of proportionality is used to control the sensitivity of the comfort reward to changes in acceleration. A higher comfort reward is obtained when $\left| {a_{i,t+1}^j - a_{i,t}}\right|$ is small, a lower comfort payoff is obtained when $\left| {a_{i,t+1}^j - a_{i,t}}\right|$ is larger.

\subsubsection{Total compensation function}

Combining the definitions of $f_{i,s}(t)$,$f_{i,e}(t)$,$f_{i,c}(t)$ above, the total reward function $U_i(t)$ can be obtained, in the case where it is known that about $W_s+ W_e+ W_c=1$, the use of a fixed value for the weights may lead to the inability of vehicle i to carry out the critical operations such as slowing down or stopping in time in face of the unexpected situation. At the same time, being too concerned about safety rewards when far away from the target vehicle may lead to insufficient driving efficiency. Therefore, it is necessary to define each weight in order to balance the three kinds of payoffs: safety, efficiency and comfort.

In order to ensure the safety of driving decisions, this paper uses the TTC value as a measure of the risk of collision with the vehicle in front of us, and uses adaptive weights $W_s(TTC)$ to define the safety reward weight values. Considering the importance of safety, the safety reward weights will be set to a higher value when the TTC value is low, so as to be more cautious and safe during driving. When the TTC value is high, the safety reward weights will be reduced so that the vehicle can be driven efficiently. Define $W_s(TTC)$ as,

\begin{align}
\begin{split}
W_s(TTC) = 
\begin{cases} 
\text{if } TTC \leq TTC_{\text{crit}}: \\
\frac{1}{3} + \frac{2}{3} \left(1 - e^{-k \left(\frac{TTC_{\text{crit}}}{TTC}\right)}\right), \\
\text{if } TTC > TTC_{\text{crit}}: \\
\frac{1}{3},
\end{cases}
\label{eq13}
\end{split}
\end{align}

where $TTC_{\text{crit}}$ is a critical TTC value, when the TTC is lower than or equal to this value, it is considered that the risk of collision may occur at this time, and the weight of safety gain is increased from 1/3 to 1. When TTC is greater than $TTC_{\text{crit}}$, it is considered to be at a safe distance, and the safety reward weight is at the lowest level 1/3. This gives $W_e, W_c$ as,

\begin{equation}
W_e = W_c = \frac{1 - W_s(TTC)}{2}
\label{eq14}
\end{equation}

with the above payoff function, this paper calculates the Nash equilibrium points in the strategy space and filters out the strategies that reach the Nash equilibrium. This results in the game reward $R_{\text{Game}}$,

\begin{equation}
R_{\text{Game}} = \frac{1}{1 + e^{-\delta (\pi - \pi_{\text{Nash}})}}
\label{eq15}
\end{equation}

where $\pi$ denotes the current strategy combination, $\pi_{\text{Nash}}$ denotes the Nash equilibrium point, and $\delta$ is a control parameter, the formula is designed to measure the proximity of the strategy combination to the Nash equilibrium and provide a reward value based on this. If more than one Nash equilibrium exists, the Nash equilibrium $\pi_{\text{Nash}}^*$ closest to the current strategy portfolio is chosen to calculate the variance,

\begin{equation}
\pi_{\text{Nash}}^* = \arg\min_{\pi_{\text{Nash}} \in \Pi_{\text{Nash}}} \|\pi - \pi_{\text{Nash}}\|
\label{eq16}
\end{equation}

where $\Pi_{\text{Nash}}$ is the set of all Nash equilibrium and $\pi_{\text{Nash}}^*$ is used in $R_{\text{Game}}$:
\begin{equation}
R_{\text{Game}} = \frac{1}{1 + e^{-\delta (\pi - \pi_{\text{Nash}}^*)}}
\label{eq17}
\end{equation}

when the strategy combination is close to the Nash equilibrium, the reward value tends to be close to 1, indicating that the strategy combination has achieved a better result in the game. On the contrary, when the strategy combination is far from the Nash equilibrium, the reward value tends to 0, indicating that the strategy combination may face a worse result in the game.

With such a reward mechanism, the system encourages participants to choose strategy combinations that are close to the Nash equilibrium in order to obtain higher rewards. Taken together, our reward function will combine factors such as safety, comfort and efficiency. Through appropriate reward and punishment mechanisms, we are able to guide the agents to exhibit safe driving behaviors and complete the task with high efficiency in the experiment. Such a reward function design will help realize a safer and more efficient transportation system and enhance the quality and feasibility of the experimental results.

\subsection*{ToM Rewards}\label{ToM}
\begin{figure*}[t]
\centering
\includegraphics[width=1\textwidth, height=0.22\textheight]{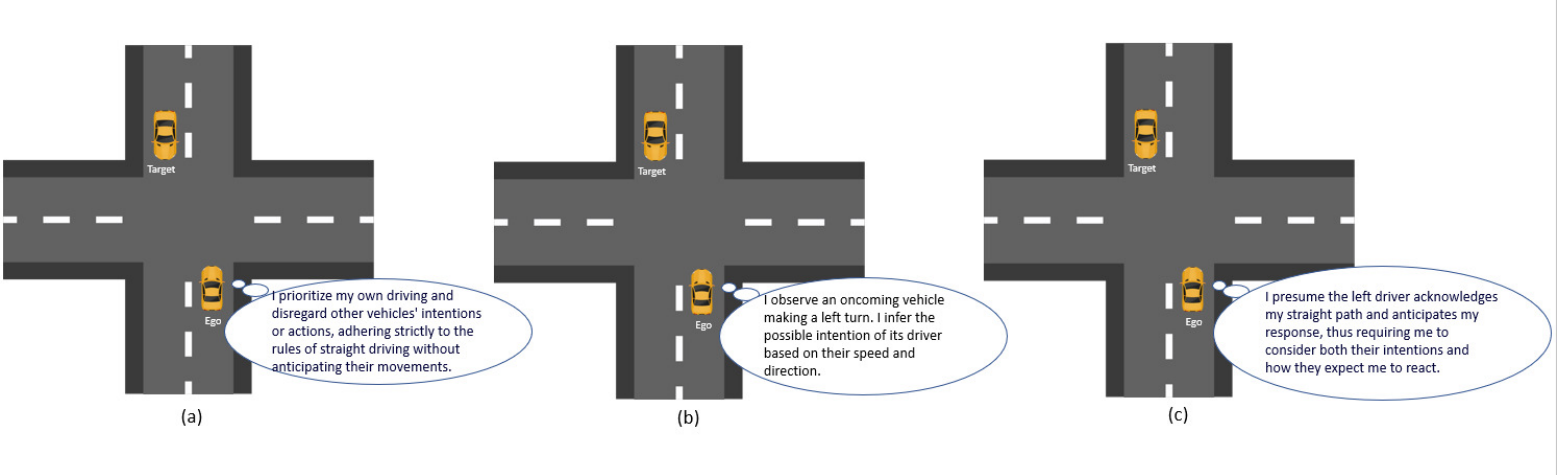}
\caption{(a) No theory of mind extrapolation (b) First-order theory of mind extrapolation (c) Second-order theory of mind}
\label{Fig2}
\end{figure*}

In order to further improve the decision-making ability of Autonomous Vehicles, this study introduces the concept of first-order Theory of Mind (ToM) inference \cite{wang2021tom2c,fuchs2021theory,tao2024enhancing} as an additional reward signal. In this context, ToM refers to the ability of human beings to understand and make inferences about their own and other people's mental states, beliefs, intentions and desires, and other internal mental processes. ToM is an important component of human social cognitive abilities. Through the Theory of Mind, people understand other people's beliefs, desires, and intentions, and see them as  the key factors in explaining other people's behaviors. ToM has different levels (orders), and, in general. ToM is categorized in first-order, second-order and higher-order inferential abilities \cite{oguntola2023theory}. First-order Theory of Mind (ToM) refers to an individual's ability to understand and infer others' mental states such as beliefs, intentions, and desires in order to predict and explain others' behaviors, as shown in Fig.\ref{Fig2}.

In this study, first-order ToM inference is applied to an unsignalized intersection scenario where there is a malicious behavior of a vehicle. By inferring the beliefs of the target vehicle, the autonomous vehicle is able to determine whether the target vehicle intends to act maliciously, such as accelerating without authorization or driving at a speed that exceeds the legal limit. Once the autonomous vehicle infers that the target vehicle has malicious behavior, it provides an additional reward signal based on this information \cite{kwon2023reward,zhao4571091multi} to enhance its own decision-making strategy.

In this study, we develop a first-order ToM model based on a Bayesian belief network for inferring the potential behaviors and beliefs of target vehicles at unsignalized intersections. The model learns and predicts the dynamic behaviors of vehicles by analyzing historical data to improve the decision-making capability of Autonomous Vehicle systems. When performing ToM inference, real-time feature data (e.g., position and speed) of the target vehicle are first input into the model using a variable elimination algorithm to perform inference and calculate the conditional probabilities of different actions taken by the target vehicle. These probabilities reflect the possible actions that the target vehicle could take given the current state.
Inferring the beliefs of the target vehicle based on the first-order ToM can be defined as $R_{ToM}$,

\begin{equation}
R_{\text{ToM}} = \sigma(-\varepsilon \cdot a_{i,t+1}^j \cdot P_{\text{Belief\_Malware}})
\label{eq18}
\end{equation}

where $\sigma$ is the sigmoid function, $\varepsilon$ is a positive moderator to control the sensitivity of the function, $P_{\text{Belief\_Malware}}$ is the probability that the target vehicle adopts a malicious behavior, and when the acceleration of the main vehicle is high, the reward is smaller if the probability of the malicious behavior of the target vehicle is also high, and vice versa if the probability of the maliciousness of the target vehicle is low, the reward is higher. When the main vehicle acceleration is low, the reward is inversely proportional to the target vehicle's malicious probability.

Combining the definitions of rewards above leads to the definition of the rewards section of the ASEQ,

\begin{equation}
R = W_1 R_{\text{ToM}} + W_2 R_{\text{Game}}
\label{eq19}
\end{equation}
where $W_1$ and $W_2$ are the weight parameters for each reward respectively, $R_{\text{ToM}}$ is the ToM reward and $R_{\text{Game}}$ is the gaming reward.

\subsection{Q-value function}
Q-learning is a reinforcement learning algorithm for learning the expected utility (i.e., Q-value) of taking each possible action in a given state. The Q-value function represents the expected payoff of taking an action in a given state. At the heart of Q-learning is the Q-value updating formula, which uses reward signals to adjust the Q-value to better estimate the long-term payoffs. The Q-value updating formula is as follows, 

\begin{align}
Q(s_{i,t}, A_{i,t}) &\leftarrow Q(s_{i,t}, A_{i,t}) + \alpha \big[ R + \gamma \max_A Q(s_{i,t+1}, A) - Q(s_{i,t}, A_{i,t}) \big]
\label{20}
\end{align}

where $Q(s_{i,t}, A_{i,t})$ is the Q value of taking action $A_{i,t}$ in state $s_{i,t}$, $R$ is the immediate reward for transferring from state $s_{i,t}$ to state $s_{i,t+1}$ computed according to II.\ref{2.D}, $\alpha$ is the learning rate, which controls the weighting of the new information during the learning process, $\gamma$ is the discount factor, which is used to balance the immediate and future rewards with the importance, and $\max_A Q(s_{i,t+1}, A)$ is the maximum Q value of all possible actions in the next state $s_{i,t+1}$, representing the best future expected reward from the next state.

\subsection{Learning and strategy selection}
In this study, we employ an $\varepsilon$-greedy strategy to guide the decision-making process of Autonomous Vehicles at unsignalized intersections. This strategy selects the best currently known action (greedy action) in most cases to take advantage of what has been learned, and also randomly selects other actions with a certain probability to explore better strategies that may be unknown. At each decision cycle, the algorithm first generates a random number between 0 and 1. If this number is greater than the current value of $\varepsilon$, the action with the highest Q value is selected to utilize the best known strategy; if it is less than or equal to $\varepsilon$, an action is randomly selected to explore possible new strategies. This mechanism helps to avoid local optimal solutions and enhances the comprehensiveness of learning.

\section{Malicious behavior simulation}
In computing the Nash equilibrium, both vehicles need to use a payoff function and corresponding weights to calculate their payoffs. However, since malicious vehicles have different behavioral characteristics, traditional payoff functions and weight settings may not properly capture their negative impact on the system. Therefore, in order to more accurately assess the behavior of malicious vehicles and protect the safety and comfort of ordinary vehicles, as well as to adapt to scenarios that include malicious behaviors, such as a left-turning vehicle that does not yield to a straight vehicle, it is necessary to adjust the payoff function for the target vehicle to ensure that all aspects of safety, efficiency, and comfort reflect the potential risks and consequences of malicious behaviors.

In this paper, the safety, efficiency, and comfort payoffs are respectively set as,

\begin{align}
f_{i,s}(t) &= e^{-\frac{d}{D}} \left(1 - e^{-\frac{TTC}{T}}\right) \\
f_{i,e}(t) &= \frac{v}{v_{\text{max}}} \\
f_{i,c}(t) &= 0
\end{align}

where $d$ is the Euclidean distance between the current vehicle and the nearest obstacle (e.g., another vehicle), $D$ is the distance decay constant to adjust the effect of distance on safety gains, and $T$ is the TTC decay constant to adjust the effect of TTC on safety gains.

\section{Simulation results and discussion}
In this section, we conduct a simulation study using Prescan/Simulink co-simulation to demonstrate the performance of the proposed decision-making solution in a two-vehicle intersection scenario. Prescan offers advanced sensor simulation and the ability to create complex traffic scenarios, making it crucial for testing and validating the performance of autonomous vehicle decision-making algorithms in variable environments. Its realistic simulation environment can model a wide range of vehicle dynamics, including malicious behavior, enabling researchers to evaluate algorithm effectiveness in a safe virtual setting. Simulink, on the other hand, provides a powerful model-based design environment that supports various engineering applications, such as control system design and automated code generation. This allows researchers to rapidly develop and test complex control strategies, facilitating their application in real vehicle testing or more advanced simulation environments, thereby accelerating the transition from theoretical research to practical applications.

\begin{figure}
\centerline{\includegraphics[width=3.5in]{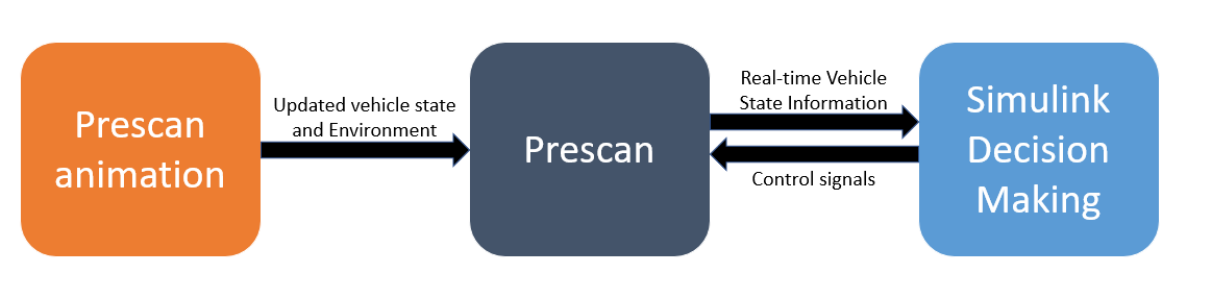}}
\caption{The structure of the simulated environment\label{fig3}}
\end{figure}

The combined use of Prescan and Simulink allows for comprehensive system testing and validation on an integrated platform, increasing the efficiency of the research and ensuring that the developed decision-making frameworks are highly useful and reliable in real-world applications. The workflow of the Prescan/Simulink joint simulation platform is shown in Fig.\ref{fig3}:

1. Data transmission: At the beginning of the simulation, Prescan first generates and sends real-time state information of the vehicle, including but not limited to position, speed, and surrounding environment data, to the Simulink decision model.

2. Decision Processes: After receiving the vehicle state information from Prescan, the Simulink decision model calculates the corresponding control signals according to the Predefined algorithms. These control signals include acceleration, steering angle, etc., which are designed to optimize the vehicle's driving strategy to cope with complex traffic situations.

3. State update: The computed control signals are transmitted back to Prescan via a virtual CAN bus, and Prescan receives these control signals to update the state of the vehicle and its position in the simulated environment. This includes adjusting the speed and direction of the vehicle, as well as updating the environment animation accordingly to reflect the new vehicle state.

4. Loop Iteration: The above process is repeated throughout the simulation period to form a closed-loop control system. In this way, the simulation platform is able to continuously simulate the dynamic response of a vehicle in a traffic environment.

\begin{figure}
\centerline{\includegraphics[width=3.5in]{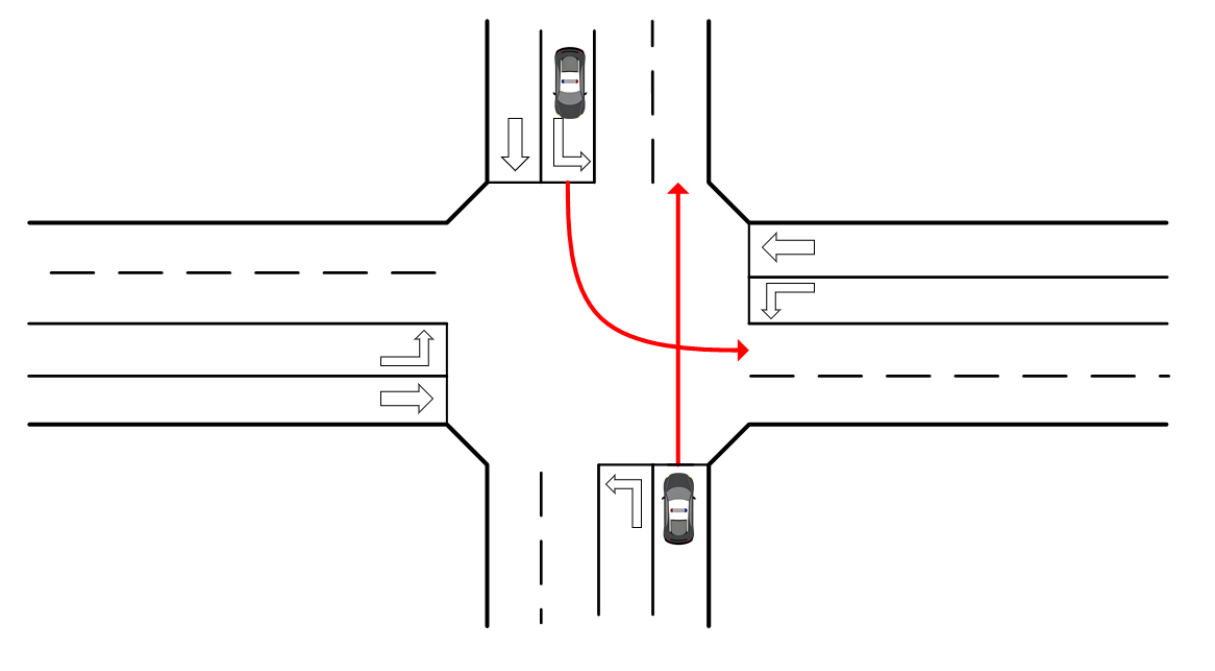}}
\caption{Schematic of vehicle behavior at unsignalized intersections\label{fig4}}
\end{figure}

\begin{figure}
\centerline{\includegraphics[width=3.5in]{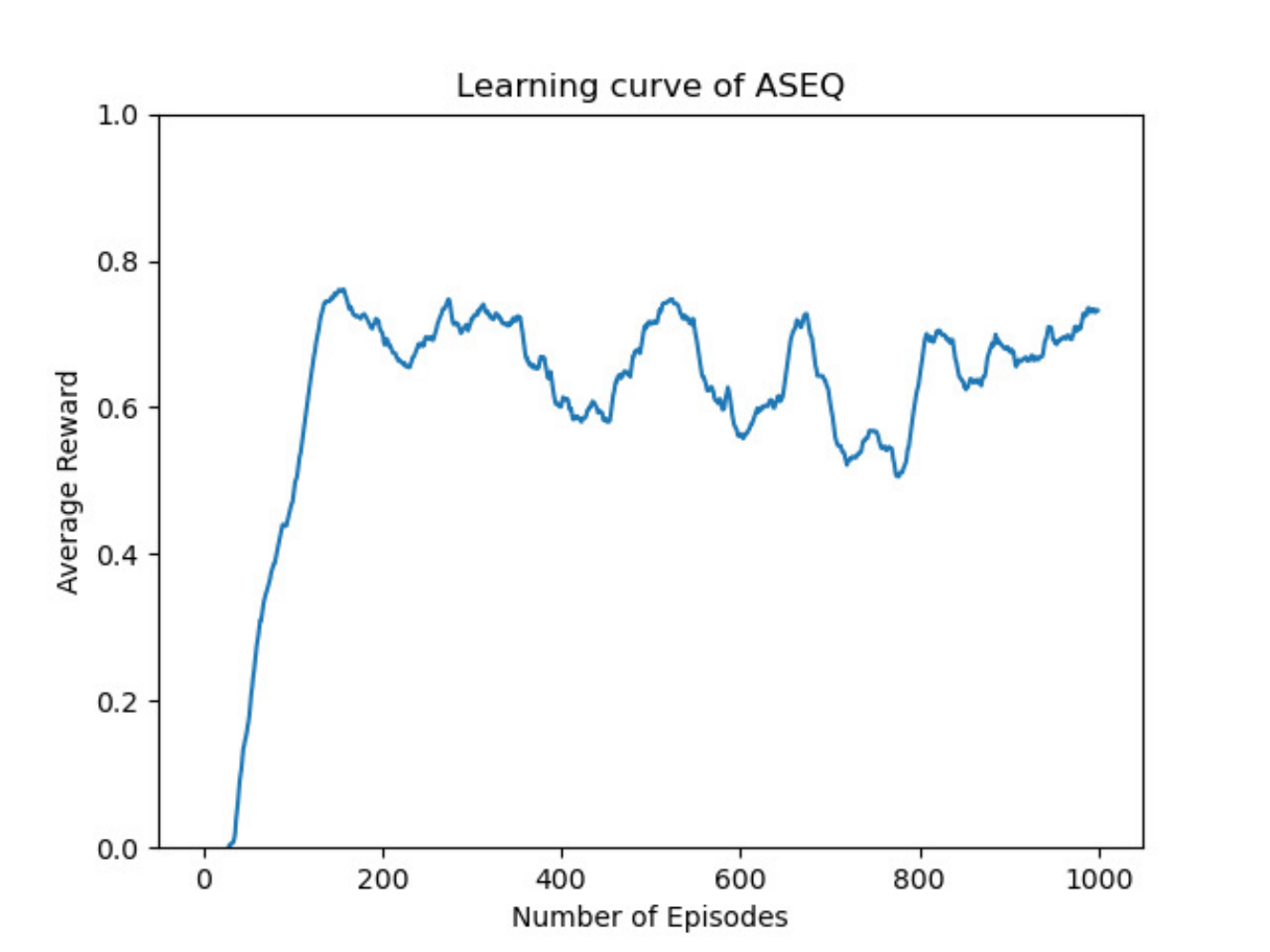}}
\caption{Learning Curve of the ASEQ\label{learn}}
\end{figure}

In order to test and verify the effectiveness of the proposed decision-making framework for dealing with complex traffic situations, the behavior patterns of malicious vehicles are especially set up. In the simulation scenario shown in Fig. \ref{fig4}, the malicious vehicle is set to need to perform a left-turn operation when it arrives at an unsignalized intersection at the same time as the ego vehicle (Autonomous Vehicles). Unlike the typical behavior of obeying traffic rules, this malicious vehicle is set to not avoid the ego vehicle that is going straight and does not perform any deceleration operation while passing through the intersection. According to the Traffic Safety Law of the People's Republic of China, the maximum prescribed speed for passing through an unsignalized intersection should be set between 30km/h and 60km/h, and in this paper, the legal maximum speed is set to 12m/s (i.e., 43.2 km/h).

Fig.\ref{learn} illustrates the training curve of ASEQ in an unsignalized intersection environment. The horizontal axis represents the number of training rounds and the vertical axis represents the average reward value. It can be seen that the average reward value gradually increases as the number of training rounds increases and stabilizes after about 200 episodes.

\subsection{Using only the game model}
\begin{figure}
\centerline{\includegraphics[width=3.75in]{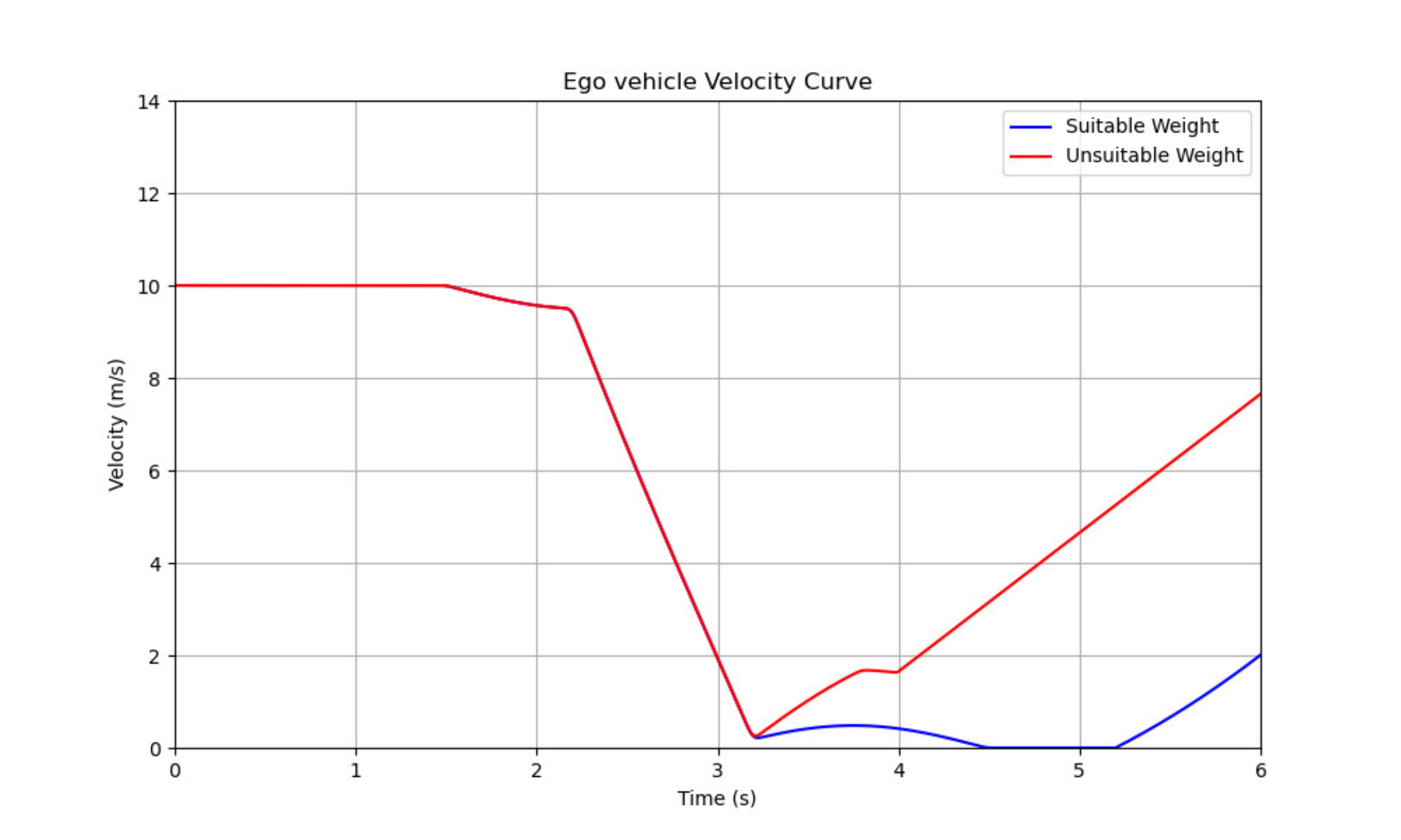}}
\caption{Case A traveling speed Curve\label{fig5}}
\end{figure}

This study first used a base game model as a comparison experiment. In this base model, the decision-making of Autonomous Vehicles relies only on traditional game theory, which takes into account interactions with other vehicles and potential conflict resolution strategies. The model mainly focuses on the trade-offs of safety, efficiency, and comfort by manually setting $W_s$, $W_e$, and $W_c$ to 0.4, 0.3, and 0.3, respectively, so that the target vehicle has an initial speed of 10\% over the speed limit, i.e., it is 13.2 m/s, and enters the intersection later than the ego vehicle; so that the ego vehicle has an initial speed of 10 m/s and can enter the intersection ahead of the target vehicle. The traveling speed curve of the ego vehicle through the intersection (0~6s) is shown in Fig.\ref{fig5}.

It can be seen that the velocity curve of the ego vehicle shows a clear process of deceleration and acceleration when using only the game model for decision-making. This behavior reflects the decision-making characteristics of the game model when dealing with complex traffic scenarios, i.e., avoiding potential conflicts by slowing down or stopping. However, between 2 and 4 s, the speed of the ego vehicle continues to decrease and reaches a minimum near 0 m/s near 4 s, indicating that the ego vehicle almost stops at the intersection. Near the minimum point of the speed, it can be seen that the acceleration of the ego vehicle undergoes a number of corrections, showing more redundant maneuvers. This strategy may result in a less efficient vehicle through the intersection and increased travel time, but overall it can satisfy the basic decision-making needs.

However, in the case of inappropriate weight settings, as shown in Fig.\ref{fig5}, the speed curve with inappropriate weight share assignment shows more sluggish decision-making, more drastic speed changes, and longer stopping times compared to the speed curve with appropriate weight share. This indicates that the inappropriate assignment of weight share in the decision-making process leads to excessive deceleration and stopping measures taken by the ego vehicle at the intersection, which significantly reduces driving efficiency.

The results highlight the importance of appropriate weight settings in the game model. Properly balanced weights can lead to smoother and more efficient decision-making, while inappropriate weights can cause excessive caution, resulting in inefficiencies. This comparison underscores the need for adaptive and context-aware weight adjustments to optimize the decision-making process in dynamic traffic environments.

\subsection{Adding adaptive weights}
\begin{figure}
\centerline{\includegraphics[width=3.75in]{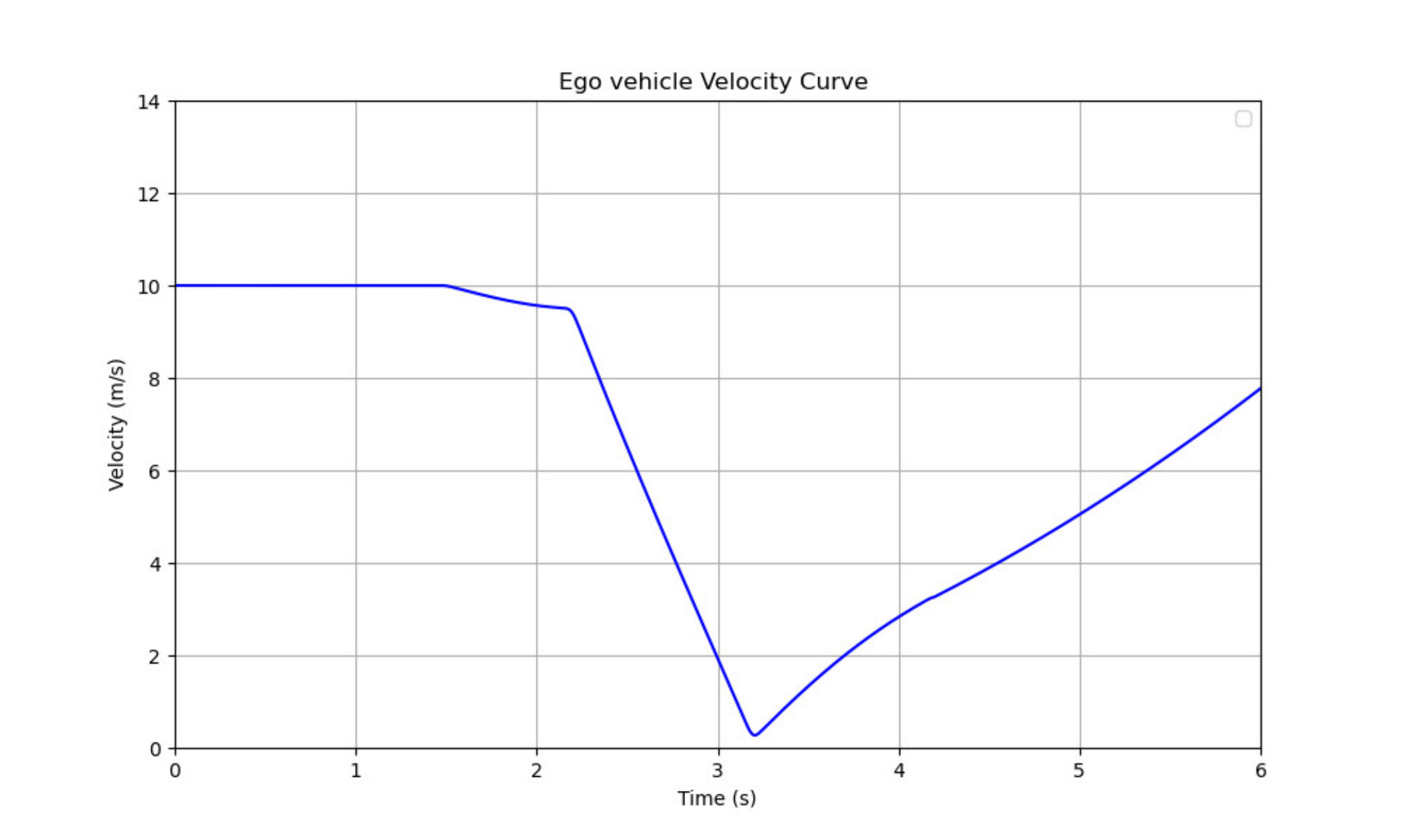}}
\caption{Case B traveling speed curve\label{fig7}}
\end{figure}

Based on A, this study adds adaptive weights with the same initial conditions as the case shown in the above figure. The traveling speed curve of the ego vehicle through the intersection (0~6s) is shown in Fig.\ref{fig7}.

By comparing with the velocity curve in A, it can be seen that the decision-making framework with the addition of adaptive weights exhibits higher efficiency and better smoothness when dealing with complex traffic scenarios. The introduction of adaptive weights enables the ego vehicle to pass through the intersection at a smoother speed while ensuring safety, which reduces unnecessary deceleration and stopping behaviors and thus improves the overall driving efficiency. In addition, near the velocity nadir, the acceleration of A undergoes multiple corrections, showing more redundant operations, while the decision-making framework with the addition of adaptive weights exhibits smoother velocity changes, reduces unnecessary operations, and improves the agility of decision-making.

This verifies the effectiveness and superiority of the adaptive weights proposed in this paper in practical applications. The adaptive weights allow the decision-making process to dynamically adjust to the traffic conditions, leading to more efficient and safer navigation through intersections. Compared to the base game model in Case A, the adaptive weights significantly reduce the instances of abrupt speed changes and stops, highlighting the importance of flexibility in weight settings for real-time decision-making.

The results demonstrate that adaptive weights can effectively balance safety, efficiency, and comfort, providing a more robust solution for autonomous vehicle decision-making in dynamic and complex traffic environments. This comparison underscores the necessity of incorporating adaptive mechanisms to enhance the performance and reliability of autonomous driving systems.

\subsection{Adding first-order ToM}
\begin{figure}
\centerline{\includegraphics[width=3.75in]{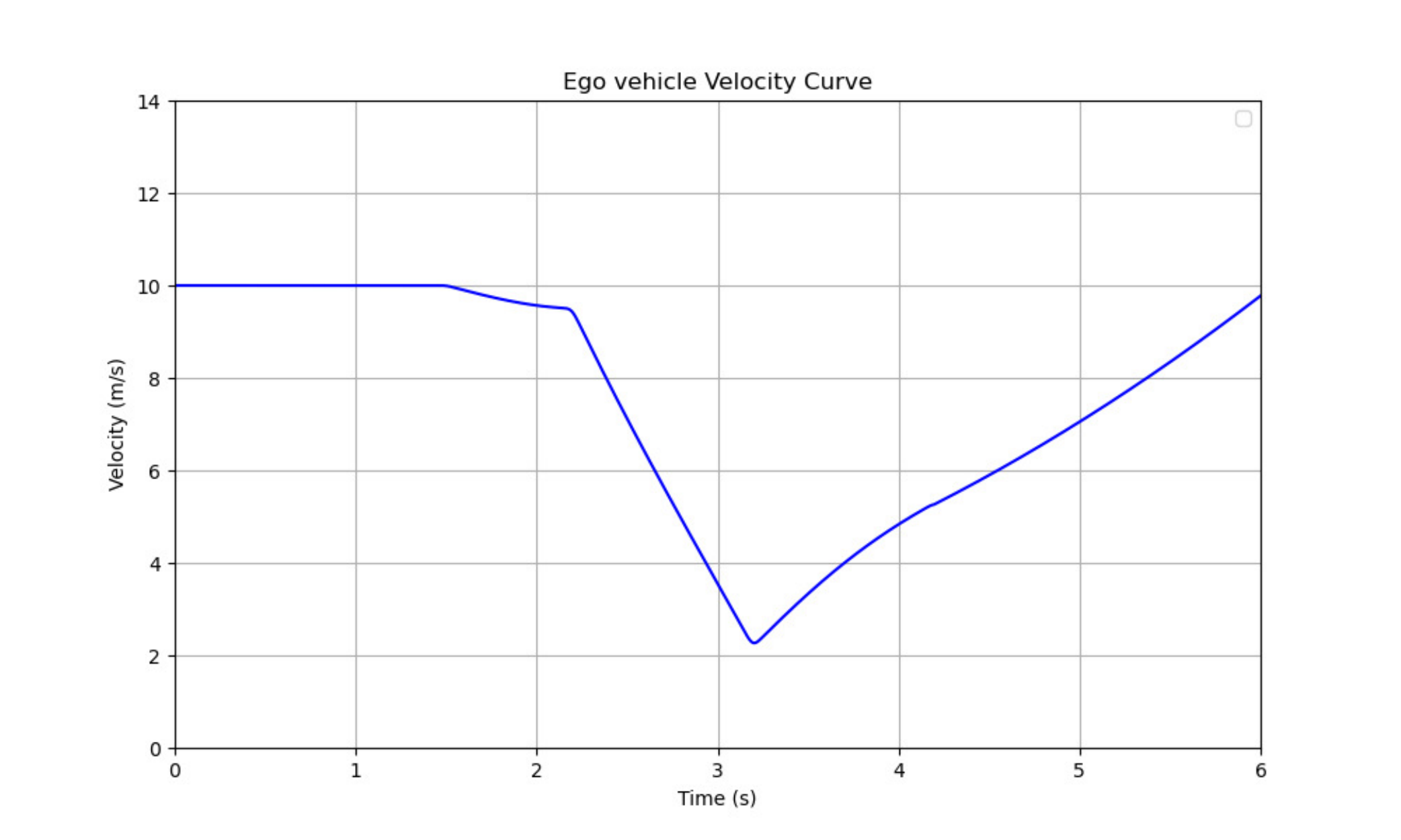}}
\caption{Case C traveling speed curve\label{fig8}}
\end{figure}

Based on B, this study adds a first-order ToM inference with the same initial conditions as the case shown in the above figure. The traveling speed curve of the ego vehicle through the intersection (0~6s) is shown in Fig.\ref{fig8}.

\begin{figure}
\centerline{\includegraphics[width=3.5in]{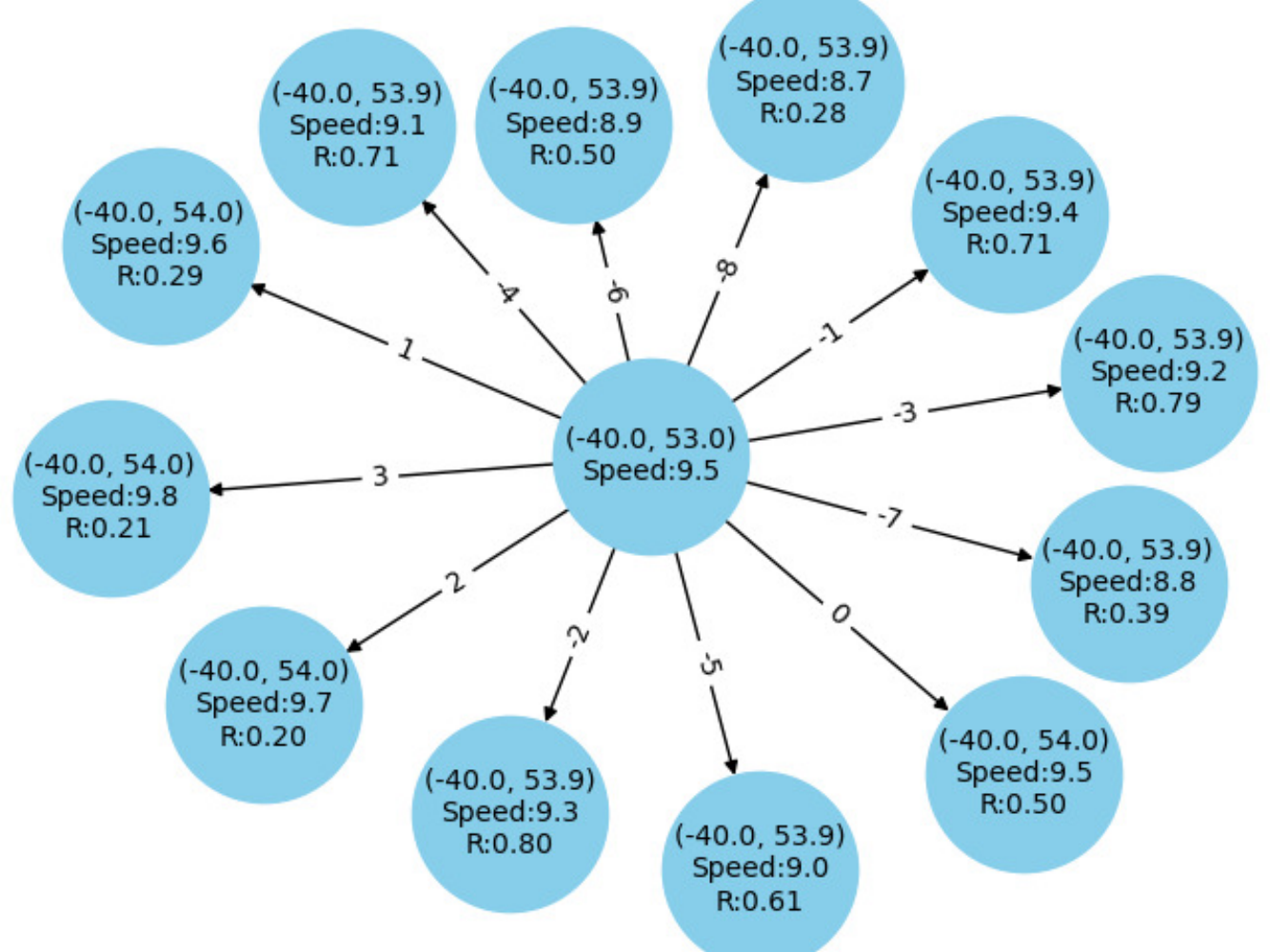}}
\caption{Action space and decision state shifts at 1.5s\label{state}}
\end{figure}

Comparison with the velocity curve in B shows that the ego vehicle has a higher minimum velocity of about 2.5 m/s in an emergency situation, compared to B. This indicates that the addition of the ToM inference makes the ego vehicle more effective in ensuring safety while maintaining a higher traveling speed and improving the overall driving efficiency.

In order to represent the decision-making process of ASEQ more concretely, we generated a figure of the decision state transfer at 1.5 seconds. As can be seen from the velocity curve figure, at 1.5 seconds, the ego vehicle has just started to reduce its speed. At this time, the Euclidean distance between the ego vehicle and the target vehicle is relatively far, so the ASEQ judges that the deceleration can be mild. Fig.\ref{state} illustrates the decision state transfer at 1.5 seconds.

As can be seen in Fig.\ref{state}, the initial states of the ego vehicle are position (-40.0, 53.0) and velocity 9.5 m/s at 1.5 s. Depending on different acceleration decisions, the states of the ego vehicle are shifted differently. As a whole, mild deceleration (e.g., acceleration of -2 $m/s^2$ and -3 $m/s^2$) is able to ensure safety while maintaining a high level of driving efficiency and comfort, and therefore receives a higher reward. In contrast, excessive deceleration or acceleration behaviors result in lower rewards, consistent with Fig.\ref{fig8}.

The results highlight the significant impact of incorporating first-order ToM inference into the decision-making framework. By predicting the intentions and beliefs of other vehicles, the ego vehicle can make more informed and proactive decisions, leading to improved safety and efficiency. Compared to B, the addition of ToM inference reduces the need for abrupt speed changes and enhances the smoothness of the vehicle's trajectory, demonstrating the value of cognitive modeling in autonomous driving systems.

\subsection{Limit case simulation}
\begin{figure}
\centerline{\includegraphics[width=3.75in]{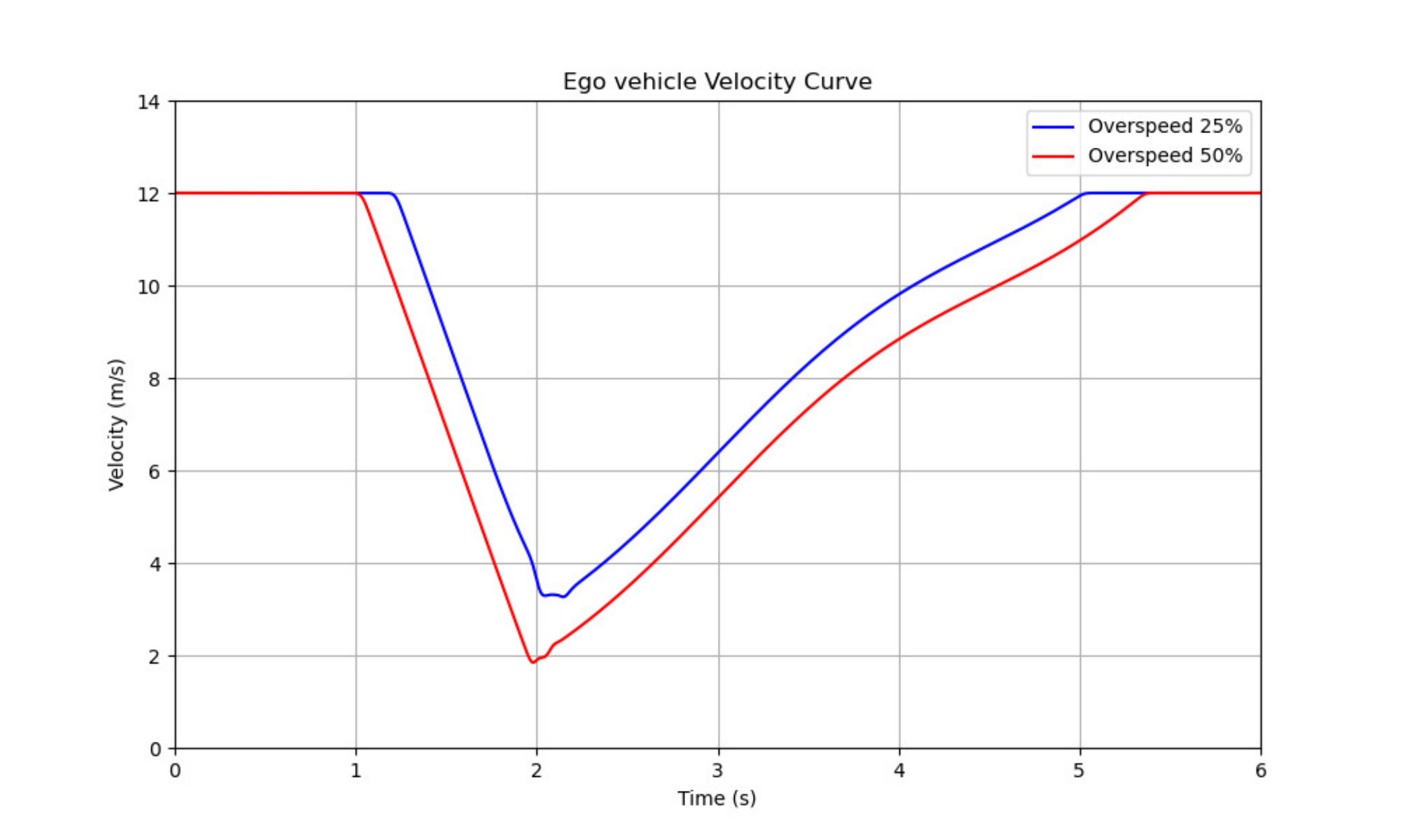}}
\caption{Case D traveling speed curve\label{fig9}}
\end{figure}

\begin{figure}
\centerline{\includegraphics[width=3.75in]{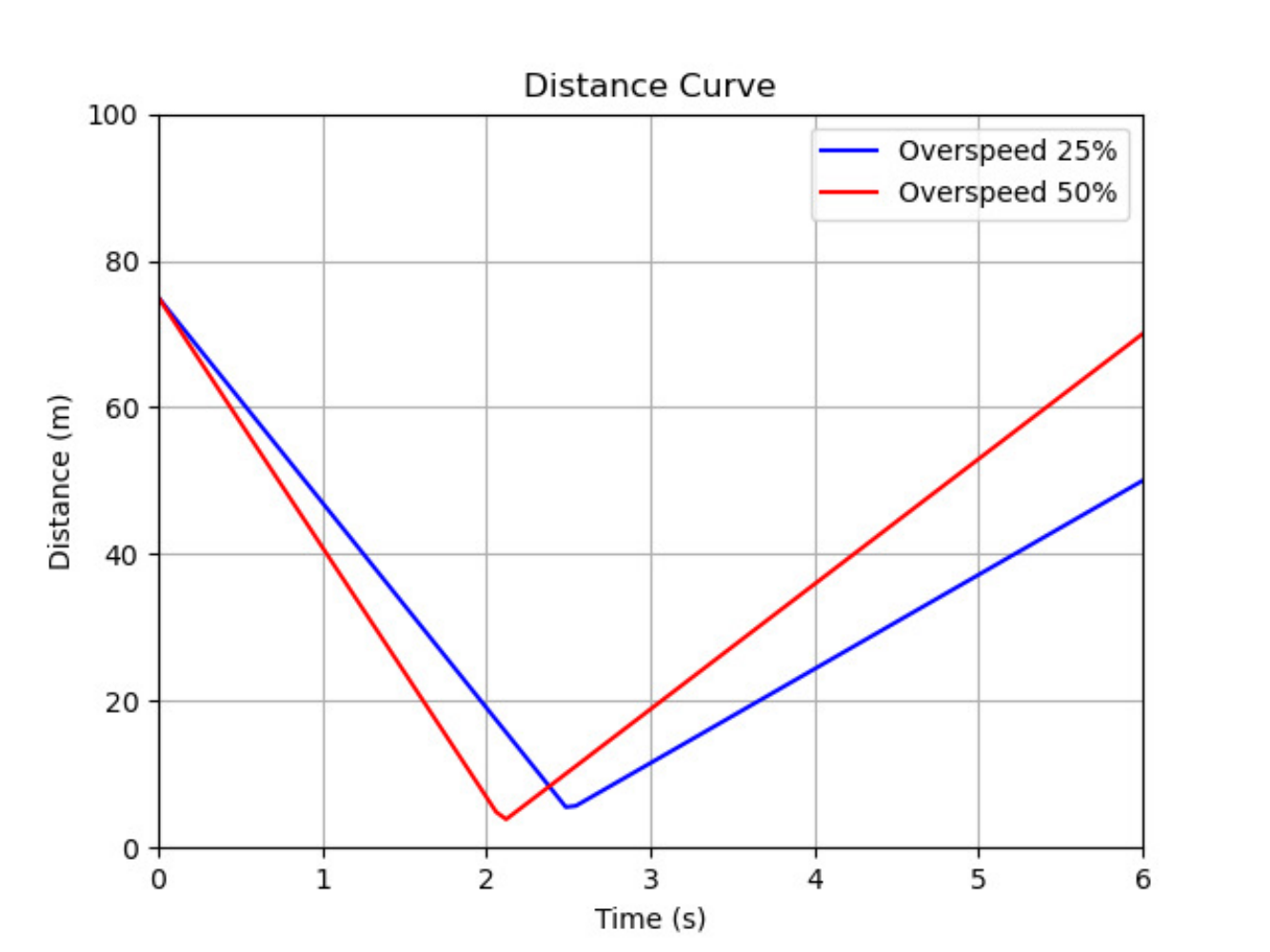}}
\caption{Case D two-vehicle Euclidean distance curve\label{fig10}}
\end{figure}

In order to test the decision-making framework for the limit case, the initial speed of the target vehicle is made to reach 25\% and 50\% over speeding, i.e. 15m/s and 18m/s respectively, and the initial speed of the ego vehicle is set to the maximum specified speed, i.e. 12m/s. The traveling speed curve (0-6s) of the speeding ego vehicle through the intersection and the Euclidean distance of the two vehicles are shown in Fig.\ref{fig9} and Fig.\ref{fig10}.

When the target vehicle exceeds the speed limit by 25\% and 50\%, the ego vehicle's decision-making framework effectively adjusts its speed to avoid collisions. In the 25\% over speeding case, the distance between the ego vehicle and the target vehicle decreases from about 70m to a minimum of about 5m, demonstrating that the ego vehicle decelerates to maintain a safe distance. In the 50\% over speeding case, the minimum distance reaches approximately 3m, indicating a more urgent need for deceleration.

The velocity curves show that the ego vehicle undergoes significant deceleration and subsequent acceleration in both scenarios, with more pronounced deceleration in the 50\% over speeding case. This behavior highlights the framework's ability to dynamically respond to varying levels of threat, ensuring safety while optimizing driving efficiency. The Euclidean distance curves further illustrate the effectiveness of the decision-making framework in maintaining safe distances even under extreme conditions.

The results from these limited case simulations underscore the robustness and adaptability of the decision-making framework. The ability to handle extreme speeding scenarios by effectively adjusting speed and maintaining safe distances highlights the framework's potential for real-world applications. The comparison between 25\% and 50\% over speeding cases demonstrates the system's capability to dynamically respond to varying levels of threat, ensuring safety while optimizing driving efficiency. This further validates the importance of incorporating ToM inference and adaptive mechanisms in enhancing the decision-making process of autonomous vehicles.

\subsection{Comparative test}
\begin{figure}
\centerline{\includegraphics[width=3.5in]{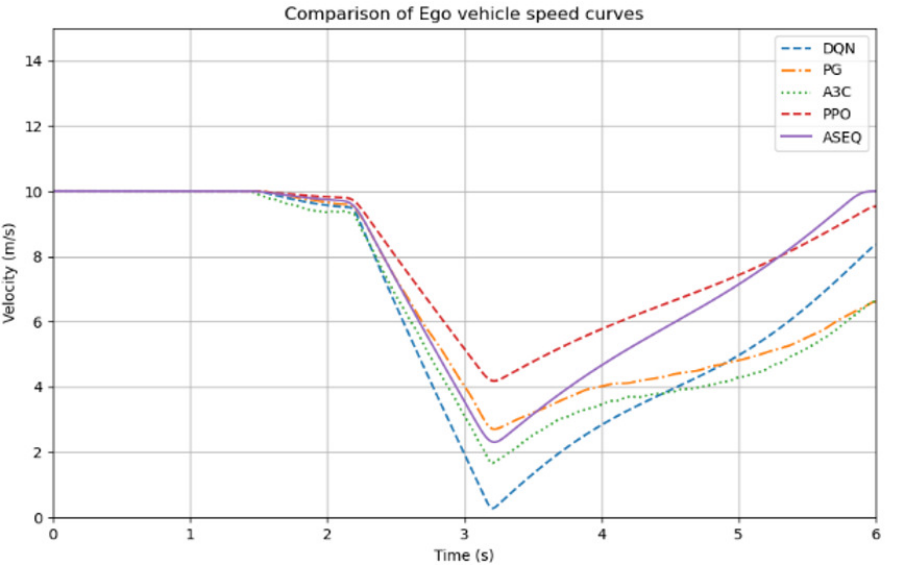}}
\caption{Speed curves for different decision-making frameworks\label{fig13}}
\end{figure}

\begin{table}[h!]
\caption{Comparison of different decision-making frameworks}
\label{table}
\centering
\setlength{\tabcolsep}{3pt}
\small 
\begin{tabular}{>{\centering\arraybackslash}m{50pt} >{\centering\arraybackslash}m{70pt} >{\centering\arraybackslash}m{70pt} >{\centering\arraybackslash}m{70pt} >{\centering\arraybackslash}m{70pt}}
\toprule
\textbf{Framework} & 
\textbf{Crossing Time(s)} & 
\textbf{Minimum Speed(m/s)} & 
\textbf{Comfort index} & 
\textbf{Minimum distance(m)} \\
\midrule
DQN & 7.208 & 0.275 & 3.728 & \underline{\textbf{7.938}} \\
PG & 7.198 & \underline{\textbf{2.603}} & 6.848 & 4.647 \\
A3C & 7.398 & 1.658 & 5.246 & 6.824 \\
PPO & 6.217 & 4.182 & 2.078 & Hit \\
ASEQ & \underline{\textbf{6.497}} & 2.306 & \underline{\textbf{2.852}} & 5.672 \\
\bottomrule
\end{tabular}
\end{table}

In the experimental analysis of this paper, we compare the performance of various baseline models (DQN, PG, A3C, PPO) with the proposed ASEQ in responding to the malicious behavior of vehicles at unsignalized intersections. The speed curve comparison graphs and various comparison metrics allow us to comprehensively assess the performance of each model in terms of safety, efficiency and comfort. From the speed curve comparison graphs, we can observe that all models take deceleration measures at some point when they detect the presence of potential malicious behaviors at unsignalized intersections. In particular, the ASEQ exhibits a significant decrease in speed after detecting the threat, but it recovers its speed the fastest, showing that it is able to quickly recover its traveling speed while ensuring safety, thus improving the overall driving efficiency. Compared to the other models, the ASEQ performs well in all comparative metrics. Although the DQN model shows strong safety, its over-conservative strategy leads to lower overall efficiency, and the PG and A3C models achieve a certain balance between safety and efficiency, but their performance in terms of comfort needs to be improved. The PPO model, although it has the strongest speed recovery ability, has obvious safety deficiencies because it fails to avoid collision effectively at critical moments.

Taken together, the ASEQ ensures absolute safety and avoids collisions by inferring the beliefs of the target vehicle through the first-order ToM and combining the safety, efficiency and comfort gains in the game rewards, especially in emergencies through the variable weighting parameter. Overall, the ASEQ performs well in all three aspects of safety, efficiency and comfort, verifying its effectiveness and superiority in dealing with malicious behaviors at unsignalized intersections.

\subsection{Generalization test}
\begin{figure}
\centerline{\includegraphics[width=3.5in]{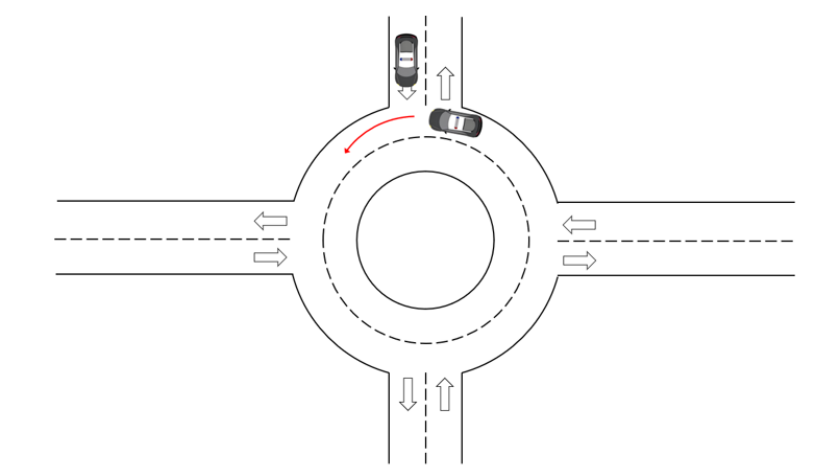}}
\caption{Schematic of vehicle behavior at unsignalized traffic circles\label{fig14}}
\end{figure}

In order to verify the generalization ability of the ASEQ framework, we apply it to a traffic circle scenario for testing. As a common traffic scenario, traffic circle has complex traffic flows and variable vehicle behaviors, as well as features of shorter reaction distance and shorter decision time, which can effectively test the adaptability and stability of the decision-making framework in different environments. Specifically, we construct a typical traffic circle scene in the simulation environment, as shown in Fig.\ref{fig14}.

\begin{figure}
\centerline{\includegraphics[width=3.5in]{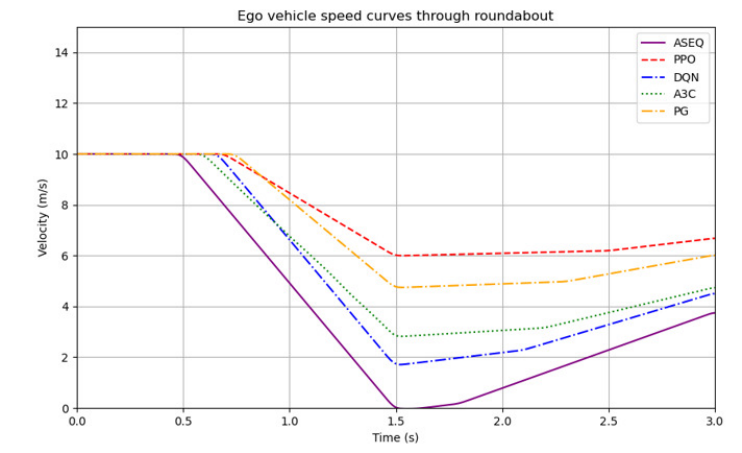}}
\caption{Speed curves for different decision-making frameworks\label{fig15}}
\end{figure}

Fig.\ref{fig15} shows the velocity curves of different decision frames during the traffic circle passage. It can be seen that the ASEQ decelerates rapidly at 0.5 seconds, with the speed dropping from 10 m/s to nearly 0 m/s, showing greater responsiveness and flexibility. This rapid deceleration capability is particularly important in emergency situations, helping to avoid potential collision risks. In contrast, the PPO and PG decelerate more slowly, with speeds remaining high at 1.5 s, presenting a greater risk of collision. In addition, DQN and A3C decelerate faster, but their speed curves fluctuate more, which may affect vehicle stability and ride comfort.

\section{Conclusion}
This study proposes an adaptive security-enhanced Q-learning framework for decision-making of Autonomous Vehicles at unsignalized intersections to deal with potentially maliciously behaving vehicles. Experimental results show that the addition of adaptive weighting and Theory of Mind (ToM) reasoning significantly improves driving efficiency, decision flexibility and safety. The excellent generalization ability of the ASEQ decision-making framework in different traffic environments is further verified through application experiments in a circular traffic scenario.

Despite the significant progress, there are still some shortcomings and future research directions. First, the types of malicious behaviors are not comprehensive enough, and future research can extend their types and complexity. Second, more complex traffic environments, such as highways and urban roads, can be considered. In addition, ToM reasoning is mainly based on first-order reasoning, and future research can explore higher-order ToM reasoning to improve the accuracy and predictability of decision-making. Another direction is to optimize the real-time and computational efficiency to ensure the performance and reliability of the algorithm in real driving environments. Finally, the experiments are mainly based on simulated data, and future research can validate the effectiveness and safety of the decision-making framework through real road tests and large-scale datasets. In conclusion, future research can further improve and validate the decision-making framework by expanding the types of malicious behaviors, exploring complex traffic environments, increasing the ToM inference order, optimizing the computational efficiency, and conducting real road tests to provide a more comprehensive and reliable solution for Autonomous Vehicles in complex traffic scenarios.

\bibliographystyle{unsrt}  
\bibliography{references}

\end{document}